\documentclass[12pt]{article}

\usepackage{newtxtext,newtxmath}

\usepackage{graphicx}

\usepackage[letterpaper,margin=1in]{geometry}

\renewenvironment{abstract}
	{\quotation}
	{\endquotation}

\date{}

\makeatletter
\renewcommand{\fnum@figure}{\textbf{Figure \thefigure}}
\renewcommand{\fnum@table}{\textbf{Table \thetable}}
\makeatother

\usepackage{scicite}

\usepackage{url}

\def\scititle{
	Understanding Artificial Neural Network's Behavior from Neuron Activation Perspective
}
\title{\bfseries \boldmath \scititle}

\author{
	Yizhou Zhang$^{1\dagger}$,
	Yang Sui$^{2}$\\
	\small$^{1}$ Amazon. Inc   \quad\quad    \small$^{2}$ Rice University.         \\
	\small$^\dagger$Corresponding author. Email: zyizhou96@gmail.com\footnote{This work is not related to the position of the corresponding author at Amazon, no matter where he affiliates with.}\and
}

\begin{document} 

\maketitle


\begin{abstract}
This paper explores the intricate behavior of deep neural networks (DNNs) through the lens of neuron activation dynamics. We propose a probabilistic framework that can analyze models' neuron activation patterns as a stochastic process, uncovering theoretical insights into neural scaling laws, such as over-parameterization and the power-law decay of loss with respect to dataset size. By deriving key mathematical relationships, we present that the number of activated neurons increases in the form of $N(1-(\frac{bN}{D+bN})^b)$, and the neuron activation should follows power-law distribution. Based on these two mathematical results, we demonstrate how DNNs maintain generalization capabilities even under over-parameterization, and we elucidate the phase transition phenomenon observed in loss curves as dataset size plotted in log-axis (i.e. the data magnitude increases linearly). Moreover, by combining the above two phenomenons and the power-law distribution of neuron activation, we derived the power-law decay of neural network's loss function as the data size scale increases. Furthermore, our analysis bridges the gap between empirical observations and theoretical underpinnings, offering experimentally testable predictions regarding parameter efficiency and model compressibility. These findings provide a foundation for understanding neural network scaling and present new directions for optimizing DNN performance.
\end{abstract}

\section{Introduction}

Deep neural networks (DNNs) have achieved unprecedented success across a wide range of applications, including natural language processing, computer vision, and scientific discovery \cite{krizhevsky2012imagenet,srivastava2014dropout,he2016deep,brown2020language}. Despite these advancements, understanding the underlying mechanisms that govern the behavior of DNNs remains a fundamental challenge. 

Recent empirical studies have revealed consistent scaling laws in neural networks, highlighting two key phenomena: over-parameterization \cite{allen2019convergence,chen2019much} and the power-law decay of loss with respect to dataset size \cite{kaplan2020scaling,hoffmann2022training} . Over-parameterization refers to the observation that DNNs with parameter sizes growing faster than dataset sizes can achieve remarkable generalization performance\cite{kaplan2020scaling}. Concurrently, the loss function of DNNs decreases approximately as a power-law function of the dataset size \cite{hoffmann2022training}. While these scaling laws have been validated empirically, theoretical explanations for their emergence remain limited and often lack generality or consistency with experimental observations.

From the generalization error perspective, classical statistical learning theory suggests that the generalization error decreases with the size of the dataset, often inversely proportional to $D$ or $\sqrt{D}$\cite{bahri2024explaining}. Some researchers attempt to extend these ideas to high-dimensional spaces by correlating the power-law exponent with the intrinsic dimensionality of the data. While this offers a plausible explanation, it falls short in explaining why such scaling laws are predominantly observed in neural networks and not in other models, such as random forests, which share similar statistical properties.

On the other hand, the capability quanta perspective, inspired by concepts from quantum mechanics, proposes that learning involves acquiring discrete units of task-solving ability, termed "capability quanta."\cite{michaud2024quantization} This theory effectively explains the observed power-law relationships but relies on counterintuitive assumptions, such as the existence of discrete generalization status, which violate the classical understanding to continuous generalization error bound, and power-law distribution of capability quanta.

In contrast, our approach focuses on the activation patterns of neurons and models their evolution as a stochastic process. By doing so, we derive theoretical results that not only explain key aspects of neural scaling laws, such as over-parameterization and the power-law decay of loss, but also align with experimental findings that are seldom discussed or newly reported. Moreover, our framework bridges the gap between empirical observations and theoretical explanations, offering new avenues for understanding the behavior of neural networks. Specifically, we analyze the evolution of neuronal activation as a stochastic process and study its relationship with dataset expansion and parameter growth. Through this approach, we derive theoretical results that align with key experimental findings:
\begin{itemize}
    \item The neural networks in over-parameterization, where the parameter size $N$ grows faster than the dataset size $D$, i.e. $N>O(D)$, can still generalize.
    \item We can observe a phase transition in the generalization process of neural network if we plot the data size in log-axis plot. This finding is implied by the figure 2 in the first neural scaling law paper from OpenAI \cite{kaplan2020scaling}, but seldom discussed by previous researches.
    \item The power-law decay of loss $L$ as a function of $D$, expressed as $L = O(D^{-s})$, where $s > 0$.
\end{itemize}

Furthermore, we extend our framework to make experimentally testable predictions regarding the role of noise-to-signal parameter ratios and the compressibility of large models. These predictions not only offer new insights into neural network scaling but also provide a foundation for future experimental validation.

The remainder of this paper is structured as follows. Section 2 presents the notations and assumptions of our probabilistic framework. In Section 3, we propose our central hypothesis, whose form is simple and clear, which assumes that when the training data size is increasing, the neuron activation distribution is steady in linear axis. In Section 4, based on our central hypothesis, we pointed out that the \textbf{neuron activation distribution is a power-law distribution} and acquired the \textbf{growth function of activated neurons}. In Section 5, based on the derived growth function, we present why neural networks under over-parameterization can still generalize, and why there exist a clear phase transition if we plot the data size in log-axis. Then using the phase transition effect and the power-law distribution, we derived the power-law decay of loss $L$ as a function of $D$. After that, in Section 6, we provides new predictions, mainly about the parameter efficiency and pruning for neural network, based on our derived conclusions and theory framework. In Section 7, we review related work on neural scaling laws and existing theoretical interpretations. Finally, Section 8 concludes with a discussion of our limitation and potential future research directions.

\section{Notations and Preliminary}
In this paper, we consider a single layer of neurons in a deep neural network. The notation used throughout are as follows:

\begin{itemize}
    \item \textbf{Number of neurons:} Let $N$ denote the total number of neurons in the layer, and the index of each neuron is denoted by $i \in \{1, 2, \dots, N\}$.
    \item \textbf{Training dataset:} The size of the training dataset is $D$.
    \item \textbf{Activation Function:} We consider a neural network that uses ReLU as activation function:
    \begin{equation}
        ReLU(x) = \begin{cases}
            x, & \text{if } x\geq0 \\
            0, & \text{if } x<0 \\
            \end{cases}
    \end{equation}
    Such a function satisfy two important properties: (1) only activated neurons contribute to the final loss of the model; and (2) only activated neurons can receive gradients in backward propagation, which means that a sample is learnt by a neuron only if the neuron is activated.
    \item \textbf{Activation of neurons:} During training on dataset $D$, not every neuron is activated by every sample (Otherwise, the neural network will lose non-linearity). Let $D_i$ represent the number of samples that activate neuron $i$. Since a neuron can only learn (i.e., be back-propagated) when it is activated, the $D_i$ also represent the sample number that neuron $i$ has learnt.
    \item \textbf{Expect of Activated Neuron Number:} Modern deep neural networks often include normalization layers (e.g., batch normalization, layer normalization, or group normalization\cite{ioffe2015batch,lei2016layer,wu2018group}), which constrain the number of neurons activated by each sample to be approximately constant across different samples, denoted as $c$. The $c$ should be only relevant to hyper-parameter setting and the neuron number $N$, and intuitively it should only increase when $N$ increases.
\end{itemize}

\section{Central Hypothesis}

In this paper, we consider a steady process that introduce a new sample into an existing dataset, i.e., moving from $D \to D+1$. This new sample is assumed to be drawn independently and identically distributed (i.i.d.) with the existing samples. Based on the above defined notations, we provide our central hypothesis:
\textbf{Stable Activation Probability Hypothesis}:
\begin{itemize}
    \item Based on the i.i.d property of the new sample, the probability of neuron $i$ being activated by the new sample is proportional to $D_i + b$, where $D_i$ is the number that neuron $i$ is activated by samples in $D$, and $b$ is a small positive constant ensuring that even unactivated neurons have a non-zero probability of being activated.
\end{itemize}

Based on the above assumption, the probability of neuron $i$ being activated by the new sample is given by:
\begin{equation}
    p_i(\sigma(i) | D_1, D_2, \dots, D_N) = \frac{cD_i + cb}{\sum_{j=1}^N (D_j + b)} = \frac{cD_i + cb}{cD + bN}=\frac{D_i + b}{D + \frac{b}{c}N}.
\end{equation}
where $b$ is a constant ensuring that unactivated neurons have a non-zero activation probability, $c$ is the approximate number of neurons activated per sample, and $cD = \sum_{j=1}^N D_j$ is the total number of activations across all neurons.

\section{The dynamic of neuron activation}
In this section, we analyze how the distribution of $D_i$ evolves as $D$ increases. Neurons are divided into two categories: \textbf{"working neurons"}, which have been activated by at least one sample, and \textbf{"free neurons"}, which have not been activated. By analyzing the stochastic process of neuron activation, we present two conclusions:
\begin{itemize}
    \item The number of working neuron, denoted as $K(D)$, increase as a function of $D$, in the form of $N(1-(\frac{bN}{cD+bN})^b)$.
    \item The distribution of activation number in the working neuron, denoted as $P(D_i=k)$, approximately follows power-law distribution: $P(D_i=k)\propto k^{-\alpha}$, where $0<\alpha<1$
\end{itemize}
\subsection{The dynamic of working neuron number}
We consider the transition from free to working neurons. Since the transition from free to working neurons is unidirectional, let $K$ denote the number of working neurons. Then, $K(D)$ is a monotonically increasing function of $D$. Based on the above probability definition, we derive the recurrence relation:
\[
K(D+1) - K(D) = \frac{(N - K(D))b}{D + \frac{b}{c}N}.
\]

By approximating the discrete difference as a continuous change, we rewrite the relation in its differential form:

\[
\frac{dK}{dD} = \frac{(N - K(D))b}{D + \frac{b}{c}N}.
\]

Solving the above differential equation yields the following solution for the number of working neurons $K(D)$:
\[
K(D) = N - \frac{(N-K_0) \left(D_0 + \frac{b}{c}N\right)^b}{\left(D + \frac{b}{c}N\right)^b},
\]
where $K_0$ and $D_0$ represent the initial conditions. Solutions can be found in the Appendix. This solution describes how the number of working neurons evolves with the growth of the dataset size $D$. For large model on large dataset (i.e. $N,D>>K_0,D_0$), $K(D)$ approximately follows the form of $N(1-(\frac{bN}{cD+bN})^b)$. 




\subsection{The distribution of working neuron's activation number}
We now consider the distribution of activation counts $D_i$ at a total sample size $D$, denoted as $P(D_i = k \mid D)$. When $D$ is sufficiently large, the distribution reaches a steady state, where $P(D_i = k \mid D) \approx P(D_i = k \mid D+1)$. 

At steady state, the clusters with $D_i = k$ have a probability proportional to $k + b$ of transitioning to $D_i = k+1$, while the clusters with $D_i = k+1$ have a probability proportional to $k+1+b$ of transitioning back to $D_i = k$. This leads to the equilibrium equation:
\[
(k + b)P(D_i = k) = (k+1 + b)P(D_i = k+1).
\]

Expanding this recurrence relation back to $P(D_i = 1)$, we find that $P(D_i = k)$ is proportional to $\frac{D_{min}+b}{k + b}P(D_i = D_{min})=\frac{\text{constant}}{k + b}$. This distribution approximates a power-law distribution $P(D_i = k)\propto k^{-\alpha}$,where $0<\alpha<1$. This range is because $\frac{k+1+b}{k+b}<\frac{k+1}{k}$, which implies $P(D_i = k)$ decrease as $k$ increases but slower than $\frac{1}{k}$. Therefore, the $P(D_i)$ forllows:
\begin{equation}
    P(D_i) = \frac{D_i^{-\alpha}}{\sum_{1}^{D_{max}}k^{-\alpha}}\approx\frac{D_i^{-\alpha}}{\int_{1}^{D_{max}}k^{-\alpha}dk}=\frac{(1-\alpha)D_i^{-\alpha}}{D_{max}^{1-\alpha}-1}\approx\frac{(1-\alpha)D_i^{-\alpha}}{D_{max}^{1-\alpha}}
\end{equation}
where $D_{max}$ is the maximum of $D_i$. Since we know $\sum D_i=cD$, we have:
\begin{equation}
    cD=\sum D_i=K(D)\mathbb{E}(D_i) \approx K(D)(1-\alpha)\frac{\int_1^{D_{max}}k^{1-\alpha}dk}{D_{max}^{1-\alpha}}=K(D)\frac{1-\alpha}{2-\alpha}\frac{D_{max}^{2-\alpha}-1}{D_{max}^{1-\alpha}}=O(K(D)D_{max})
\end{equation}
Thus, we have $D_{max}=O(\frac{cD}{K(D)})$.

\section{Understanding Deep Neural Network Behavior}
In this section, we apply the above two conclusions to understand three key empirical phenomenons:
\begin{itemize}
    \item The generalization of over-parametrized neural network
    \item The phase transition of neural network generalization
    \item The power-law decay of neural network loss regarding the increase of data size scale.
\end{itemize}

\subsection{The generalization of over-parametrized neural network}

From the above results, we can understand why overparamterized neural network (i.e. the neural network with $N>O(D)$) highly generalize. Given that every sample activates $c$ neurons, we consider the average sample number of every base learner (working neuron), which is defined as:
\[
\frac{cD}{K(D)}=\frac{c}{\frac{N}{D}(1-(\frac{D_0+\frac{b}{c}N}{D+\frac{b}{c}N})^b)+\frac{K_0}{D}(\frac{D_0+\frac{b}{c}N}{D+\frac{b}{c}N})^b}
\]

Suppose that N grows with D with linearity and $N,D >> D_0$, i.e. N=kD=O(D), then the $\frac{cD}{K(D)}$ will become:
\[
\frac{cD}{K(D)}=\frac{c}{C_1+\frac{1}{D+C_2}}\approx O(c)
\]
which grows when D and N grow. Thus, it means that if N grows with D with linearity, then the average sample of each base learner will grow together with D, finally leading to underfitting in some settings. Thus, N should grow a little bit faster then D, which aligns with experimental observation that $N^{0.74}\propto D$.

\subsection{The phase transition of neural network generalization}

\begin{figure}
    \centering
    \includegraphics[width=0.4\linewidth]{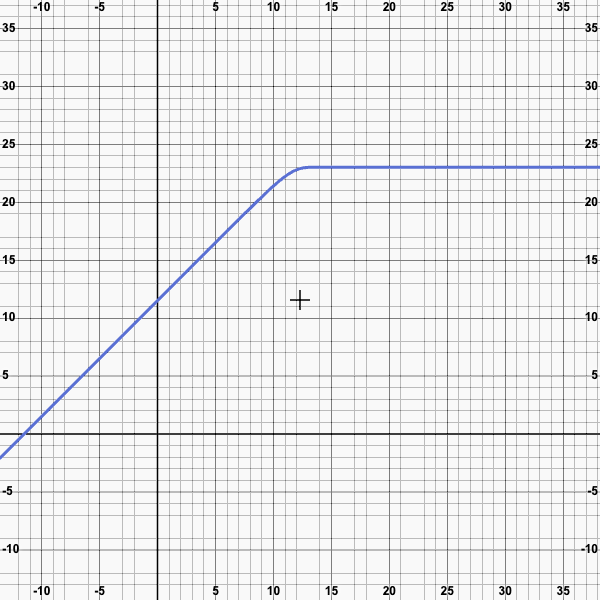}
    \caption{The relationship between $\log(K(D))$ (y-axis) and $\log(D)$ (x-axis). As we can see, when $0<\log(D)<10$, which means that $D$ crosses multiple magnitudes, the $\log(K(D))$ fits in the linear relationship very well. And after $\log(D)>11$, which means that $D$ crosses the magnitude of $\log(\frac{b}{c}N)$, the $\log(K(D))$ fits in the constant approximation very well}
    \label{fig:approximation}
\end{figure}

In \cite{kaplan2020scaling}, particularly the figure 2, the authors reported the curves about how neural language model's performance change with data size $D$ when the parameter size is given. They reported the curves of model with various parameter sizes. And it can be observed that the validation loss of all models shows a phase transition phenomenon. More specifically, at very beginning, the model's loss is very steady, which looks approximately as a flat line. Then as data increases, the model will get into a transition phase, in which the model's loss dramatically reduces. After that, the model's loss again become a approximately steady flat line.

Such phenomenon is very counterintuitive for machine learning area, in which the generalization error bound of a model should be a continuous function that decrease as data size increase \cite{germain2016pac}. However, in this section, we are going to explain why such phase transition process generally happens for different models.

Given a series of neurons, $K(D)$ satisfies the following approximation:

\[
\log(K(D)) =
\begin{cases}
\text{constant}, & \text{if } \log(D) \geq \log(\frac{b}{c}N)+\epsilon \\
\log(D)+\text{constant}, & \text{if } \log(D) \leq \log(\frac{b}{c}N)-\epsilon \\
\text{Inapproximable} & \text{otherwise}
\end{cases}
\]

We provide the theoretic analysis to the above approximation. To give more intuitive justification, we put a figure (Fig. \ref{fig:approximation}) showing how $\log(K(D))$ change with $\log(D)$ when $N=10^{11}$ (i.e. 10B, the normal scale of large language models), $c=10^5$ (approximately the square root of $N$) and $b=10^3$. From the figure, we can see that across multiple magnitudes that is lower than $\log \frac{b}{c}N$, the linear approximation fits very well. And when the magnitude crosses $\log \frac{b}{c}N$, it dramatically change to a constant. 

When the $\log(D)<\log(\frac{b}{c}N)-\epsilon$, i.e. $cD\ll bN$, we see $\log(K(D))=\log(D)+\text{constant}$, which implies $K(D)=O(D)\ll N'$. This means that the working neuron number much fewer than free neurons, implying the output of the whole network is dominant by a random noise generated by a free noise with initial parameters. Thus, the loss should be close to the loss of a randomly initialized neural network, denoted as $L_{noise}$. Meanwhile, when $\log(D)>\log(\frac{b}{c}N)+\epsilon$, i.e., $bN\ll cD$, from the over-parameterization nature of deep neural network, we can know that the sample numbers is sufficiently enough for the network to learn parameters that are close to optimal. We denote this loss as $L_{opt}$. It is obviously that there exists a clear gap between $L_{opt}$ and $L_{noise}$. Thus, we can observe a phase transition phenomenon when we apply $\log$ axis  

\subsection{The neural scaling law}
In this section, we will discuss how to approximately derive neural scaling law by analyzing the above dynamics and distributions that we acquired. 

Since the activation probabilities of the "working neurons" in our theory also follows power-law distribution, it is very natural to relate the "working neuron" with the 'ability quanta' in \cite{michaud2024quantization}. Meanwhile, when $bN\ll cD$ (i.e., $\log(D)>\log(\frac{b}{c}N)+\epsilon$), the probability that a free neuron is activated is close to 0:
\begin{equation}
    K(D+1) - K(D) = \frac{(N - K(D))b}{D + \frac{b}{c}N}\leq\frac{bN}{D}\approx0.
\end{equation}
Thus, when $D$ is sufficiently large, the overall loss of the neural network is only contributed by the "working neurons" that are activated. In other words, the overall loss can be represented by the expect of the loss contributed by the activated neuron:
\begin{equation}
\begin{aligned}
    L(D) &= \mathbb{E}_iL_i(D_i) = \sum_i^{K(D)}p_iL_i(D_i) = \sum_i^{K(D)}p(\sigma(i) | D_1, D_2, \dots, D_N)L_i(D_i) \\
   & = \sum_i^{K(D)}\frac{D_i + b}{D + \frac{b}{c}N}L_i(D_i)\approx \sum_i^{K(D)}\frac{D_i}{D}L_i(D_i)
\end{aligned}
\end{equation}

where $i$ represent the neuron index and $p_i = p(\sigma(i) | D_1, D_2, \dots, D_N)$ is the probability that the $i$-th neuron is activated. 

Since a neuron's output is actually the output of the sub-network that consists of its predecessor neurons and itself, then each working neuron's loss should also satisfy the phase transition phenomenon that we previously discussed, i.e. :
\[
\log(K(D_i)) =
\begin{cases}
\text{constant}, & \text{if } \log(D_i) \geq \log(\frac{b_i}{c_i}N_i)+\epsilon \\
\log(D_i)+\text{constant}, & \text{if } \log(D_i) \leq \log(\frac{b_i}{c_i}N_i)-\epsilon \\
\text{Inapproximable} & \text{otherwise}
\end{cases}
\]

Due to the symmetry of neural network architecture, for two neuron's in the same layer, their predecessor layers are the same, and thus share same configuration of $b_i,c_i,N_i$. These same configuration also implies same random initialized loss $L'_{noise}$ and optimal loss $L'_{opt}$ of these sub-networks, which implies:
\[
L_i =
\begin{cases}
L'_{opt}, & \text{if } \log(D_i) \geq \log(\frac{b_i}{c_i}N_i)+\epsilon \\
L'_{noise}, & \text{if } \log(D_i) \leq \log(\frac{b_i}{c_i}N_i)-\epsilon \\
L_t(D_i) & \text{otherwise}
\end{cases}
\]
where $L_t(D_i)$ means the transition phase loss. We denoted the shared configuration as $b',c',N'$. Therefore, we have:

\begin{equation}
\begin{aligned}
    L(D) = \mathbb{E}_iL_i(D_i) = & L'_{noise}P(\log{D_i}<\log(\log(\frac{b'}{c'}N')-\epsilon)\\
    &+L'_{opt}P(\log{D_i}>\log(\log(\frac{b'}{c'}N')+\epsilon)\\
    &+(\mathbb{E}L_{t})P(\log(\frac{b'}{c'}N')-\epsilon<\log{D_i}<\log(\log(\frac{b'}{c'}N')+\epsilon)
\end{aligned}
\end{equation}
where $\mathbb{E}L_{t}$ refers to the expect of loss from those neurons in transition phase. 

Meanwhile, as we derived previously in Section 4, the distribution of $D_i$ follows $D_i^{-\alpha}$:
\begin{equation}
    P(D_i) = \frac{(1-\alpha)D_i^{-\alpha}}{D_{max}^{1-\alpha}}
\end{equation}
where $0<\alpha<1$ is the shape parameter of $D_i$'s distribution. 

Then we have:
\begin{equation}
    P(\log(\frac{b'N'}{c'})-\epsilon<\log{D_i}<\log(\frac{b'N'}{c'})+\epsilon)=\frac{(\frac{b'N'}{c'})^{1-\alpha}(e^\epsilon-e^{-\epsilon})}{D_{max}^{1-\alpha}}=\frac{(b'N')^{1-\alpha}(e^\epsilon-e^{-\epsilon})}{c'D_{max}^{1-\alpha}}
\end{equation}

Since $\epsilon\rightarrow0$, we can do Taylor expansion to $e^\epsilon$ to acquire $e^{\epsilon}-e^{-\epsilon}=2\epsilon$. Thus $P(\log(\frac{b'N'}{c'})-\epsilon<\log{D_i}<\log(\frac{b'N'}{c'})+\epsilon)=O(\frac{b'N'}{c'D_{max}}\epsilon)$. When $\epsilon\rightarrow0$ and $b'N'\ll c'D_{max}$, we have $P(\log(\frac{b'N'}{c'})-\epsilon<\log{D_i}<\log(\frac{b'N'}{c'})+\epsilon)\rightarrow0$

Therefore, we have:
\begin{equation}
\begin{aligned}
    L(D) &= \mathbb{E}_iL_i(D_i) \approx  L'_{noise}P(\log{D_i}<\log(\log(\frac{b'}{c'}N')-\epsilon)+L'_{opt}P(\log{D_i}>\log(\log(\frac{b'}{c'}N')+\epsilon)\\
    &\approx L'_{opt}+(L'_{noise}-L'_{opt})P(\log{D_i}<\log(\log(\frac{b'}{c'}N')-\epsilon)\\
    &\approx L'_{opt}+(L'_{noise}-L'_{opt})P(\log{D_i}<\log(\log(\frac{b'}{c'}N'))\\
    &=L'_{opt}+(L'_{noise}-L'_{opt})\frac{(b'N')^{1-\alpha}-1}{{c'D_{max}}^{1-\alpha}}
\end{aligned}
\end{equation}
From the expression of $K(D)$, we know that $K(D)$ monotonically increases and $K(D)<N$. Thus, when $D$ is sufficiently large, $K(D)=O(1)$. Thus, $D_{max} = O(cD)$ have:
\begin{equation}
\begin{aligned}
    L(D)
    &\approx L'_{opt}+(L'_{noise}-L'_{opt})\frac{(b'N')^{1-\alpha}-1}{(c'\cdot cD)^{1-\alpha}}
\end{aligned}
\end{equation}
As we discussed, $0<\alpha<1$. Thus, the $L(D)$ is actually a power-law function of $D$ and $N$

\section{Predictions of Unknown Phenomena}

In the previous sections, we demonstrated how the proposed model explains two key phenomena in scaling laws: over-parameterization and the power-law decay of loss with dataset size. In this section, we extend the analysis to predict some previously unexplored phenomena based on our model.

We focus on the noise-to-signal parameter ratio, which quantifies the relationship between the model's total parameters $N$ and the scale of free neurons $N - K(D)$. Particularly, this ratio can help us understand the optimal or sub-optimal ratio of pruning and compressing a neural network \cite{li2017pruningfiltersefficientconvnets,hassibi1992second}, which shows significance in many areas \cite{li2023q}. Using the formula for $K(D)$, we can derive:
\[
\frac{N - K(D)}{N} = \frac{(bN)^b}{(cD + bN)^b}=\frac{(\frac{b}{c}N)^b}{(D + \frac{b}{c}N)^b}.
\]
When $N$ grows faster than $D$, this ratio approaches 1 as $N$ increases. This result implies that the proportion of ineffective parameters in a large model increases as both the model size $N$. When N grows faster than linearity of $D$, we also see that when datasize increases, the ineffective parameter ratio increases too. Based on this observation, we propose several predictions:

1. Larger models are inherently more compressible.

2. With the same architecture, models trained on larger datasets achieve a lower compression ratio compared to those trained on smaller datasets. The ratio should approximately decrease as a power-low function of $D$ in the form of $(\frac{C_1}{D+C_1})^{C_2}$

3. With the same dataset, larger models generally achieve higher compression ratios than smaller models. The ratio should approximately grows as a power-low function in the form of $(\frac{N^\alpha}{C_1+N^\alpha})^{C_2}$

4. Over time, naively pre-trained large models will exhibit lower capability density compared to smaller models trained with more data.

It is worth noting that the second conclusion has been empirically verified in recent work. Specifically, the concept of capability density, introduced in \cite{xiao2024densing}, highlights that the effective parameter size of a model relative to its actual parameter size is increasing over time. This aligns with our theoretical prediction that the density of useful parameters in larger, naively pre-trained models diminishes relative to smaller, efficiently trained models. The study shows that the capability density of open-source models has doubled approximately every three months, supporting the idea that smaller, well-trained models can achieve comparable or even superior performance with fewer parameters.

\section{Related Work}

This work builds upon and extends previous studies in neural scaling laws \cite{kaplan2020scaling,hoffmann2022training}, which examine the relationship between compute, parameter size, dataset size, and the loss of neural networks. Since compute can be expressed as a function of parameter size and dataset size, we primarily focus on the relationships between parameter size $N$ and dataset size $D$, as well as the relationship between dataset size $D$ and neural network loss $L$.

Here, we define:
\begin{itemize}
    \item \textbf{Dataset size $D$}: The number of training samples used to optimize the neural network.
    \item \textbf{Parameter size $N$}: The total number of trainable parameters in the neural network.
    \item \textbf{Loss $L$}: The objective function value that the neural network minimizes during training, often representing the prediction error on the dataset.
\end{itemize}

\subsection{Generalization under Overparameterization}

One critical aspect of neural scaling laws is the phenomenon of overparameterization. As neural networks scale, the parameter size $N$ increases faster than the dataset size $D$. Mathematically, this can be expressed as:
\[
N > O(D),
\]
indicating that the parameter size grows superlinearly with respect to the dataset size. Empirically, \( N^{0.74} \propto D \)\cite{kaplan2020scaling}, indicating that the parameter size \( N \) scales sub-linearly with respect to the dataset size \( D \). This overparameterization has been shown to improve generalization, despite the apparent risk of overfitting. Recent studies have empirically and theoretically validated this relationship across various architectures and tasks, highlighting that sufficiently large models are capable of capturing complex patterns in data, even with finite datasets.

\subsection{Power-Law Relationship between Loss and Dataset Size}

Another key observation in neural scaling laws \cite{kaplan2020scaling} is the power-law decay of loss $L$ with increasing dataset size $D$. Specifically, the loss can be expressed as:
\[
L(D) = O\left(D^{-s}\right),
\]
where $s > 0$ is a positive constant. This power-law relationship has been consistently observed in both language and vision models, suggesting that increasing the dataset size leads to diminishing returns in reducing loss. The exponent $s$ depends on the model architecture, the quality of the data, and the task, but its existence reflects a universal trend in neural scaling.

The insights from these two aspects of neural scaling laws provide a foundation for the theoretical analysis in this paper, particularly the derivation of scaling behavior in loss and the role of overparameterization in improving model performance.

\subsection{Interpretations of Neural Scaling Laws}

Existing works attempting to interpret neural scaling laws generally follow two primary approaches.

\textbf{1. Generalization Error Perspective.}  
From the perspective of classical statistical learning theory, the generalization error is typically inversely proportional to either the dataset size $D$ or the square root of $D$ \cite{lotfi2022pac,germain2016pac}. Based on this, some researchers have proposed that the power-law exponent observed in scaling laws could be explained as the intrinsic dimensionality of the data in high-dimensional spaces. However, this explanation encounters a critical limitation: theoretically, it should also apply to other machine learning models, such as random forests. In practice, however, the power-law relationships are predominantly observed in neural networks and not in other machine learning models. This discrepancy casts doubt on the universality of this approach.

\textbf{2. Capability Quanta Perspective.}  
Another approach, inspired by ideas from quantum mechanics, has been proposed by researchers with a background in physics \cite{michaud2024quantization}. This approach hypothesizes the existence of "capability quanta," analogous to energy quanta in quantum mechanics. Each capability quantum represents a discrete unit of task-solving ability. When a model acquires a new capability quantum, it unlocks a specific ability. Under this framework, the distribution of capability quanta follows a power-law distribution, and the data and parameter requirements for learning each quantum also exhibit power-law growth. 

While this theory successfully explains the observed power-law relationships between loss, dataset size, and parameter size, it has several significant limitations:
\begin{itemize}
    \item \textbf{Counterintuitive Assumptions:} The theory assumes the existence of capability quanta. These quanta has \textbf{discrete} status, i.e. "learnt" and 'non-learnt'. And the losses contributed by quantas with different status have a clear gap. As admitted by the authors themselves in their discussion of paper limitation, this assumption is very counterintuitive, and violates the observation that neural networks are optimized gradually. In machine learning area, the generalization error bound of a model is usually characterized as a continuous function of sample number $D$.
    \item \textbf{Specific Distribution Assumptions:} It further assumes that the distribution of capability quanta follows a power-law distribution. Other distributions, such as exponential or Poisson distributions with appropriate parameters, are not considered, making the framework appear overly tailored to fit the power-law observations.
\end{itemize}

These limitations highlight the challenges in fully explaining neural scaling laws and suggest the need for more robust and empirically consistent theoretical frameworks.

In this paper, we propose a new perspective to understand the behavior of neural networks. By analyzing the activation patterns of neurons and modeling their evolution as a stochastic process, we study how neural networks scale with increasing dataset size $D$ and parameter size $N$. Through this approach, we derive conclusions consistent with existing experimental results, including the benefits of over-parameterization and the power-law decay of loss with $D$. Furthermore, within this theoretical framework, we provide experimentally testable predictions.

\section{Conclusion, Limitation and Future Work}
This paper provides a probabilistic framework for understanding the impact of individual neuron activation patterns on the behavior of deep neural networks. The derived results align with experimental observations and offer testable predictions that can guide future research. 

Concurrently, this paper leaves huge space for future works to improve on, including but not limited to:
\begin{itemize}
    \item Our stable activation probability hypothesis implies a data sample clustering process that is similar as Dirichlet Process and Pitman-Yor process \cite{teh2010dirichlet,sato2010topic,teh2009indian}. A natural question is: what is the deep mechanism of such clustering process in neural network?
    \item In our paper, we mainly discuss the power-law decreasing relationship between $D$ and loss. Deriving the relationship about $N$ needs to analyze the $L_{opt}$ term. This term is not related to our data size, but the expressive power of a neural network with \textbf{limited parameters}.
    \item Extending this analysis to multiple layers. Our current analysis did not specify the configuration. As a result, we do not need to specify which layer we are analyzing. Therefore, the form of power-law relationship that we acquired can be extended to every layers, explaining the dynamics of deep neural networks. However, since different layers have different hyper-parameter configurations, precisely calculate the aggregation of them to acquire the precise equation for the whole neural network is still non-trivial. We suggest such analysis may require background knowledge in fractal theory and recursive analysis.
    \item Why is shape unimportant for Transformer? This phenomenon, which is also important in neural scaling law, is still not answered. Intuitively, understanding this phenomenon requires us to precisely calculate the aggregation of different layer's configurations so that we can acquire the precise equation for the whole neural network.
\end{itemize}

\bibliographystyle{plain}
\bibliography{science_template}

\begin{thebibliography}{10}

\bibitem{allen2019convergence}
Zeyuan Allen-Zhu, Yuanzhi Li, and Zhao Song.
\newblock A convergence theory for deep learning via over-parameterization.
\newblock In {\em International conference on machine learning}, pages 242--252. PMLR, 2019.

\bibitem{bahri2024explaining}
Yasaman Bahri, Ethan Dyer, Jared Kaplan, Jaehoon Lee, and Utkarsh Sharma.
\newblock Explaining neural scaling laws.
\newblock {\em Proceedings of the National Academy of Sciences}, 121(27):e2311878121, 2024.

\bibitem{brown2020language}
Tom Brown, Benjamin Mann, Nick Ryder, Melanie Subbiah, Jared~D Kaplan, Prafulla Dhariwal, Arvind Neelakantan, Pranav Shyam, Girish Sastry, Amanda Askell, et~al.
\newblock Language models are few-shot learners.
\newblock {\em Advances in neural information processing systems}, 33:1877--1901, 2020.

\bibitem{chen2019much}
Zixiang Chen, Yuan Cao, Difan Zou, and Quanquan Gu.
\newblock How much over-parameterization is sufficient to learn deep relu networks?
\newblock {\em arXiv preprint arXiv:1911.12360}, 2019.

\bibitem{germain2016pac}
Pascal Germain, Francis Bach, Alexandre Lacoste, and Simon Lacoste-Julien.
\newblock Pac-bayesian theory meets bayesian inference.
\newblock {\em Advances in Neural Information Processing Systems}, 29, 2016.

\bibitem{hassibi1992second}
Babak Hassibi and David Stork.
\newblock Second order derivatives for network pruning: Optimal brain surgeon.
\newblock {\em Advances in neural information processing systems}, 5, 1992.

\bibitem{he2016deep}
Kaiming He, Xiangyu Zhang, Shaoqing Ren, and Jian Sun.
\newblock Deep residual learning for image recognition.
\newblock In {\em Proceedings of the IEEE conference on computer vision and pattern recognition}, pages 770--778, 2016.

\bibitem{hoffmann2022training}
Jordan Hoffmann, Sebastian Borgeaud, Arthur Mensch, Elena Buchatskaya, Trevor Cai, Eliza Rutherford, Diego de~Las Casas, Lisa~Anne Hendricks, Johannes Welbl, Aidan Clark, et~al.
\newblock Training compute-optimal large language models.
\newblock {\em arXiv preprint arXiv:2203.15556}, 2022.

\bibitem{ioffe2015batch}
Sergey Ioffe.
\newblock Batch normalization: Accelerating deep network training by reducing internal covariate shift.
\newblock {\em arXiv preprint arXiv:1502.03167}, 2015.

\bibitem{kaplan2020scaling}
Jared Kaplan, Sam McCandlish, Tom Henighan, Tom~B Brown, Benjamin Chess, Rewon Child, Scott Gray, Alec Radford, Jeffrey Wu, and Dario Amodei.
\newblock Scaling laws for neural language models.
\newblock {\em arXiv preprint arXiv:2001.08361}, 2020.

\bibitem{krizhevsky2012imagenet}
Alex Krizhevsky, Ilya Sutskever, and Geoffrey~E Hinton.
\newblock Imagenet classification with deep convolutional neural networks.
\newblock In {\em Advances in neural information processing systems}, pages 1097--1105, 2012.

\bibitem{lei2016layer}
Jimmy Lei~Ba, Jamie~Ryan Kiros, and Geoffrey~E Hinton.
\newblock Layer normalization.
\newblock {\em ArXiv e-prints}, pages arXiv--1607, 2016.

\bibitem{li2017pruningfiltersefficientconvnets}
Hao Li, Asim Kadav, Igor Durdanovic, Hanan Samet, and Hans~Peter Graf.
\newblock Pruning filters for efficient convnets, 2017.

\bibitem{li2023q}
Xiuyu Li, Yijiang Liu, Long Lian, Huanrui Yang, Zhen Dong, Daniel Kang, Shanghang Zhang, and Kurt Keutzer.
\newblock Q-diffusion: Quantizing diffusion models.
\newblock In {\em Proceedings of the IEEE/CVF International Conference on Computer Vision}, pages 17535--17545, 2023.

\bibitem{lotfi2022pac}
Sanae Lotfi, Marc Finzi, Sanyam Kapoor, Andres Potapczynski, Micah Goldblum, and Andrew~G Wilson.
\newblock Pac-bayes compression bounds so tight that they can explain generalization.
\newblock {\em Advances in Neural Information Processing Systems}, 35:31459--31473, 2022.

\bibitem{michaud2024quantization}
Eric Michaud, Ziming Liu, Uzay Girit, and Max Tegmark.
\newblock The quantization model of neural scaling.
\newblock {\em Advances in Neural Information Processing Systems}, 36, 2024.

\bibitem{sato2010topic}
Issei Sato and Hiroshi Nakagawa.
\newblock Topic models with power-law using pitman-yor process.
\newblock In {\em Proceedings of the 16th ACM SIGKDD international conference on Knowledge discovery and data mining}, pages 673--682, 2010.

\bibitem{srivastava2014dropout}
Nitish Srivastava, Geoffrey Hinton, Alex Krizhevsky, Ilya Sutskever, and Ruslan Salakhutdinov.
\newblock Dropout: a simple way to prevent neural networks from overfitting.
\newblock {\em The journal of machine learning research}, 15(1):1929--1958, 2014.

\bibitem{teh2009indian}
Yee Teh and Dilan Gorur.
\newblock Indian buffet processes with power-law behavior.
\newblock {\em Advances in neural information processing systems}, 22, 2009.

\bibitem{teh2010dirichlet}
Yee~Whye Teh et~al.
\newblock Dirichlet process.
\newblock {\em Encyclopedia of machine learning}, 1063:280--287, 2010.

\bibitem{wu2018group}
Yuxin Wu and Kaiming He.
\newblock Group normalization.
\newblock In {\em Proceedings of the European conference on computer vision (ECCV)}, pages 3--19, 2018.

\bibitem{xiao2024densing}
Chaojun Xiao, Jie Cai, Weilin Zhao, Guoyang Zeng, Xu~Han, Zhiyuan Liu, and Maosong Sun.
\newblock Densing law of llms.
\newblock {\em arXiv preprint arXiv:2412.04315}, 2024.

\end{thebibliography}
\clearpage

\appendix

\section{Derivative Equation Solution}

\textbf{Step 1: Rewrite the Equation}
First, rewrite the equation in the standard form of a first-order linear differential equation:

\[
\frac{dK}{dD} + \frac{b}{D + \frac{b}{c}N} K = \frac{bN}{D + \frac{b}{c}N}
\]

Here:
\begin{itemize}
    \item \( K \) is a function of \( D \).
    \item \( N \), \( b \), and \( c \) are constants.
\end{itemize}

\textbf{Step 2: Determine the Integrating Factor}

The standard form of a first-order linear differential equation is:

\[
\frac{dK}{dD} + P(D) K = Q(D)
\]

Where:
\begin{itemize}
    \item \( P(D) = \frac{b}{D + \frac{b}{c}N} \)
    \item \( Q(D) = \frac{bN}{D + \frac{b}{c}N} \)
\end{itemize}

The integrating factor \( \mu(D) \) is calculated as:

\[
\mu(D) = e^{\int P(D) \, dD} = e^{\int \frac{b}{D + \frac{b}{c}N} \, dD} = \left(D + \frac{b}{c}N\right)^b
\]

\textbf{Step 3: Apply the Integrating Factor}

Multiply both sides of the differential equation by the integrating factor:

\[
\mu(D) \frac{dK}{dD} + \mu(D) P(D) K = \mu(D) Q(D)
\]

This simplifies to:

\[
\frac{d}{dD} \left[ \mu(D) K(D) \right] = \mu(D) Q(D)
\]

Substituting the known \( \mu(D) \) and \( Q(D) \):

\[
\frac{d}{dD} \left[ \left(D + \frac{b}{c}N\right)^b K(D) \right] = \frac{bN}{D + \frac{b}{c}N} \left(D + \frac{b}{c}N\right)^b
\]

Simplify the right-hand side:

\[
\frac{d}{dD} \left[ \left(D + \frac{b}{c}N\right)^b K(D) \right] = bN \left(D + \frac{b}{c}N\right)^{b-1}
\]

\textbf{Step 4: Integrate Both Sides}

Integrate both sides with respect to \( D \):

\[
\left(D + \frac{b}{c}N\right)^b K(D) = N \left(D + \frac{b}{c}N\right)^b + C
\]

Where \( C \) is the constant of integration.

\textbf{Step 5: Solve for \( K(D) \)}

Divide both sides by \( \left(D + \frac{b}{c}N\right)^b \):

\[
K(D) = N + \frac{C}{\left(D + \frac{b}{c}N\right)^b}
\]

\textbf{General Solution}

Therefore, the general solution to the differential equation is:

\[
K(D) = N + \frac{C}{\left(D + \frac{b}{c}N\right)^b}
\]

Where \( C \) is an arbitrary constant determined by initial conditions.

\textbf{Example}

Suppose there is an initial condition \( K(D_0) = K_0 \), then:

\[
K_0 = N + \frac{C}{\left(D_0 + \frac{b}{c}N\right)^b}
\]

Solving for \( C \):

\[
C = \left(K_0 - N\right) \left(D_0 + \frac{b}{c}N\right)^b
\]

Substituting \( C \) back into the general solution gives the particular solution that satisfies the initial condition:

\[
K(D) = N + \frac{(K_0 - N) \left(D_0 + \frac{b}{c}N\right)^b}{\left(D + \frac{b}{c}N\right)^b}
\]

This provides the complete solution to the differential equation.

\section{Approximation of $\log(K(D))$}
Here, we justify the following approximation:
\[
\log(K(D)) =
\begin{cases}
\text{constant}, & \text{if } \log(D) \geq \log(\frac{b}{c}N)+\epsilon \\
\log(D)+\text{constant}, & \text{if } \log(D) \leq \log(\frac{b}{c}N)-\epsilon \\
\text{Inapproximable} & \text{otherwise}
\end{cases}
\]

Let us define $x = \log(D)$. Then we can rewrite $\log(K(D))$ as a function of $x$:
\begin{equation}
    \log(K(D))=f(x)=\log\left(N \left(1 - \left(\frac{bN}{ce^x + bN}\right)^b\right)\right)
\end{equation}

When \( ce^x \) is much smaller than \( bN \) (i.e., \( ce^x \ll bN \)), the term \( \frac{bN}{ce^x + bN} \) approaches 1. This simplification allows us to approximate the function and its derivative as follows:

\[
\frac{bN}{ce^x + bN} \approx 1 - \frac{ce^x}{bN}
\]

Raising both sides to the power of \( b \), we get:

\[
\left(\frac{bN}{ce^x + bN}\right)^b \approx \left(1 - \frac{ce^x}{bN}\right)^b \approx 1 - b \cdot \frac{ce^x}{bN} = 1 - \frac{ce^x}{N}
\]

Substituting this approximation back into the original function \( f(x) \):

\[
f(x) = \ln\left(N \left(1 - \left(\frac{bN}{ce^x + bN}\right)^b\right)\right) \approx \ln\left(N \cdot \frac{ce^x}{N}\right) = \ln(ce^x) = \ln c + x
\]

Taking the derivative of the approximated function:

\[
f'(x) \approx \frac{d}{dx} (\ln c + x) = 1
\]

Alternatively, when \( ce^x \) is much larger than \( bN \) (i.e., \(bN \ll  ce^x \)), we can approximate the function and its derivative as follows:

\[
\frac{bN}{ce^x + bN} \approx \frac{bN}{ce^x}
\]

Similarly, we substitute this approximation back into the original function \( f(x) \):

\[
f(x) = \log\left(N \left(1 - \left(\frac{bN}{ce^x + bN}\right)^b\right)\right) \approx \log N + b\log\left( 1-\left(\frac{bN}{ce^x}\right)^b\right)
\]

Since the magnitude of $ce^x$ is larger than $bN$, we have $\left(\frac{bN}{ce^x}\right)^b\rightarrow0$. Thus, we have:
\[
f(x)  \approx \log N + b\log\left( 1-0\right) = \log N
\]
which is a constant.

\end{document}